\pdfoutput=1
\PassOptionsToPackage{numbers,sort&compress}{natbib}
\documentclass{article}

\usepackage[preprint]{neurips_2025}

\usepackage[utf8]{inputenc} 
\usepackage[T1]{fontenc}    
\usepackage{hyperref}       
\usepackage{url}            
\usepackage{booktabs}       
\usepackage{amsfonts}       
\usepackage{amsthm}         
\usepackage{nicefrac}       
\usepackage{microtype}      
\usepackage{xcolor}         
\usepackage{graphicx}
\usepackage{amsmath,amssymb}
\usepackage{makecell}
\usepackage{multirow}
\usepackage{mathtools}
\usepackage[normalem]{ulem}

\theoremstyle{plain}
\newtheorem{theorem}{Theorem}
\newtheorem{theorem*}{Theorem*}
\newtheorem{definition}{Definition}
\newtheorem{assumption}{Assumption}
\newtheorem{lemma}{Lemma}

\title{UC-MOA: Utility-Conditioned Multi-Objective Alignment for Distributional Pareto-Optimality}

\author{
  Zelei Cheng\thanks{Equal contribution.} \\
    Northwestern University\\
    Evanston, USA
  \And
  Xin-Qiang Cai\footnotemark[1] \\
    RIKEN-AIP\\
    Tokyo, Japan
  \AND
  Yuting Tang \\
    The University of Tokyo\\
    Tokyo, Japan
  \And
  Pushi Zhang \\
    Microsoft Research Asia\\
    Beijing, China
  \And
  Boming Yang \\
    The University of Tokyo\\
    Tokyo, Japan
  \AND
  Masashi Sugiyama \\
    RIKEN-AIP \&\\ The University of Tokyo\\
    Tokyo, Japan
  \And
  Xinyu Xing \\
    Northwestern University\\
    Evanston, USA
}

\begin{document}

\maketitle

\begin{abstract}
Aligning Large Language Models (LLMs) to individual user needs is paramount for effective personalization. However, capturing the diverse and often nuanced preferences of different users presents a significant challenge. Existing approaches may not adequately represent complex preference trade-offs, can be computationally expensive, or rely on the LLM's limited numerical reasoning capabilities for personalization. We introduce Utility-Conditioned Multi-Objective Alignment (UC-MOA), a novel framework for personalized LLM alignment. UC-MOA develops a diverse set of utility functions, each representing a distinct preference structure, to cater to a wide spectrum of user requirements. These utility functions are constructed with a theoretical guarantee to ensure the resulting set of preference-aligned policies can achieve distributional Pareto-optimality. Instead of conditioning on raw numerical values, UC-MOA employs a ``utility-conditioned'' method, using a symbolic token that represents an optimal utility function index. This allows the LLM to learn an implicit mapping between the token and the desired personalized response characteristics through fine-tuning, thereby enabling effective personalization. We conduct a user study demonstrating UC-MOA's superior ability to satisfy diverse user preferences compared to baseline methods, and perform quantitative experiments showing that UC-MOA achieves superior Pareto fronts and maintains high computational efficiency, offering a scalable and robust solution for multi-objective personalized LLM alignment.
\end{abstract}

\section{Introduction}
\label{sec:intro}
Large Language Models (LLMs) have demonstrated impressive capabilities across a wide range of natural language tasks and increasingly serve as powerful tools for real-world applications~\cite{ouyang2022training,bai2022training,stiennon2020learning}. 
Developing personalized LLMs are essential for business monetization, as it enables tailored user experiences that enhance engagement, retention, and revenue~\cite{lyu2024llm, zhang2024personalization, li2024can}.
However, achieving effective personalization is non-trivial, as it requires addressing diverse and often conflicting user preferences, which cannot be adequately represented by a single scalar reward \cite{cheng2023everyone, lee2024aligning}.
Conventional fine-tuning approaches, such as Proximal Policy Optimization (PPO) \cite{schulman2017proximal} and Direct Preference Optimization (DPO) \cite{rafailov2024direct} optimize a single or aggregated reward signal, which fails to capture the multi-objective nature of real-world user requests. Consequently, there is an urgent need for multi-objective alignment methods that can balance multiple goals to satisfy diverse individual preferences.

Existing methods for multi-objective alignment in LLMs face challenges when applied to fine-grained personalization. Traditional Multi-Objective Reinforcement Learning from Human Feedback (MORLHF) uses linear scalarization of reward signals~\cite{li2020deep}. This approach struggles to capture complex, non-linear preference structures unique to individual users and typically requires training separate models for different preference weightings, making it computationally prohibitive for wide-scale personalization. 
More flexible techniques like Rewarded Soups~\cite{rame2024rewarded} interpolate between multiple fine-tuned models, but still incur the high cost of training and storing multiple LLMs.
A recent alternative, Reward-in-Context (RiC)~\cite{yangrewards} seeks to achieve personalization within a single model by embedding numerical reward targets directly into the prompt, thereby guiding the LLM toward a user-specified point in the reward space. 
However, a significant drawback of RiC 
is their reliance on a continuous reward space. To achieve satisfactory personalization performance, the LLM must possess robust numerical reasoning capabilities~\cite{yang2024number}. Specifically, it needs to accurately judge which training examples or potential generations are quantitatively closer to the user's multi-dimensional preference vector. 
This dependence on precise numerical interpretation can be a bottleneck, as LLMs often struggle with such fine-grained quantitative discernment~\citep{levy-geva-2025-language}, potentially leading to inconsistent or suboptimal personalization outcomes.

To address these limitations and advance personalized LLM alignment, we propose Utility-Conditioned Multi-Objective Alignment (UC-MOA). At the core of UC-MOA is the development of a diverse set of non-linear utility functions, each intended to capture a distinct user preference structure across multi-dimensional reward outcomes. 
Importantly, this diverse set of utility functions is constructed with theoretical guarantees ensuring that the policies aligned with these utilities can achieve distributional Pareto-optimality~\cite{cai2023distributional}. This means the framework can identify policies that offer optimal trade-offs across objectives, considering the entire distribution of outcomes, without being dominated by other achievable policies.
Instead of injecting numerical reward targets, which can tax the LLM's numerical reasoning, we employ a symbolic conditioning mechanism. For each response, we determine which utility function in our diverse set it best satisfies. We then associate this response with a discrete symbolic token, \texttt{<max\_utility\_index>} $i^*$, where $i^*$ is the index of that optimal utility function. Through fine-tuning, the LLM learns an implicit mapping between this symbolic token and the nuanced characteristics of responses that align with the corresponding underlying utility function. This strategy enables fine-grained personalization according to diverse preference profiles while sidestepping the LLM’s numerical reasoning limitations. 



To summarize, our contributions are:
\begin{itemize}
    \item UC-MOA introduces a novel method for personalized LLM alignment by training and leveraging a diverse set of utility functions to simulate varied user preferences. These utility functions are designed to ensure the resulting set of aligned policies can achieve distributional Pareto-optimality.
    \item We propose a symbolic utility-conditioned mechanism where the LLM learns a mapping from utility indices to desired response styles, enabling robust personalization while avoiding the pitfalls of direct numerical reward injection.
    \item Our comprehensive experiments demonstrate the superiority of UC-MOA in satisfying user preference, achieving competitive Pareto fronts for diverse preference profiles, 
    and maintaining high computational efficiency, offering a scalable and effective solution for personalizing LLMs to complex, multi-faceted user preferences.
\end{itemize}

\section{Related Work}

\subsection{Multi-Objective Personalization of LLMs}

Traditional LLM alignment, using methods such as PPO~\citep{christiano2017deep, ouyang2022training, bai2022training} and DPO~\cite{rafailov2024direct}, primarily optimizes for a \emph{single} reward signal. This is ill-suited to the realities of personalized assistants, where users care about several, often conflicting, attributes simultaneously~\cite{kirk2024benefits}. To address this, MORLHF~\cite{li2020deep} extends PPO for multiple objectives, often by treating user preferences as different weighting schemes. However, its common reliance on linear scalarization struggles with complex, non-linear preference trade-offs, and scaling to many unique user profiles by training separate models is impractical. Rewarded Soups~\cite{rame2024rewarded} offer parameter-merging strategies but still face scalability challenges for fine-grained, diverse personalization.

More direct conditioning approaches aim to achieve personalization within a single model.
RiC~\cite{yangrewards} attempts to condition the LLM directly on reward preferences by embedding target reward values into the input prompt.
For instance, the model may be instructed to generate responses with specified numerical scores for attributes such as helpfulness and harmlessness. 
However, the effectiveness of RiC relies heavily on the LLM’s capacity for numerical reasoning, particularly in accurately interpreting continuous input signals, which can present a significant limitation for reliable personalization.

Our UC-MOA framework offers a distinct approach to personalized multi-objective alignment. 
It begins by learning a library of strictly non-decreasing, non-linear utility functions, each capturing a unique preference structure. 
The monotonicity constraint ensures that increases in any individual reward dimension do not decrease the overall utility, thereby preserving consistency with rational user behavior.
Rather than conditioning the model on numerical reward targets, UC-MOA uses symbolic tokens representing the selected utility functions to guide a single fine-tuned model. 
This circumvents the LLM's numerical reasoning limitations while still enabling fine-grained personalization across diverse multi-objective preferences.

\subsection{Distributional Pareto-Optimality}

Effective multi-objective personalization requires more than optimizing expected reward values, as users often have preferences concerning the distribution of outcomes (e.g., risk, consistency) across objectives~\cite{zhang2021policy, bai2022training}. Distributional Pareto-Optimal Multi-Objective RL (DPMORL)~\citep{cai2023distributional} provides a formal grounding for distributional Pareto-optimality. DPMORL defines Pareto optimality over return distributions, where a policy is optimal if no other policy stochastically dominates it. Critically, DPMORL shows that policies maximizing any strictly increasing utility function over the multi-objective return distribution are part of the Distributional Pareto set. This links diverse utility functions to the exploration of distributionally optimal trade-offs.

UC-MOA leverages this by constructing a diverse set of utility functions designed to span a comprehensive range of preference structures over distributional outcomes. By conditioning the LLM on symbolic indices tied to these utility functions, UC-MOA implicitly guides the model towards policies that are distributionally Pareto-optimal. This ensures that the achieved personalization is not only diverse but also adheres to a rigorous standard of optimality regarding the full spectrum of possible outcome distributions, without requiring the LLM to perform explicit distributional calculations.

\section{Proposed Technique}
\label{sec:proposed}

Effectively personalizing LLMs to diverse user preferences across multiple objectives is a significant challenge. While recent methods such as RiC~\citep{yang2024rewards} attempt personalization by appending raw numerical reward values directly to prompts, LLMs often struggle with precise numerical inputs, limiting their ability to reliably adapt to nuanced user needs, as discussed in Section~\ref{sec:intro}. Our proposed framework, UC-MOA, circumvents this limitation to achieve robust multi-objective personalization.

\begin{figure*}[t]
\centering
\includegraphics[width=\textwidth]{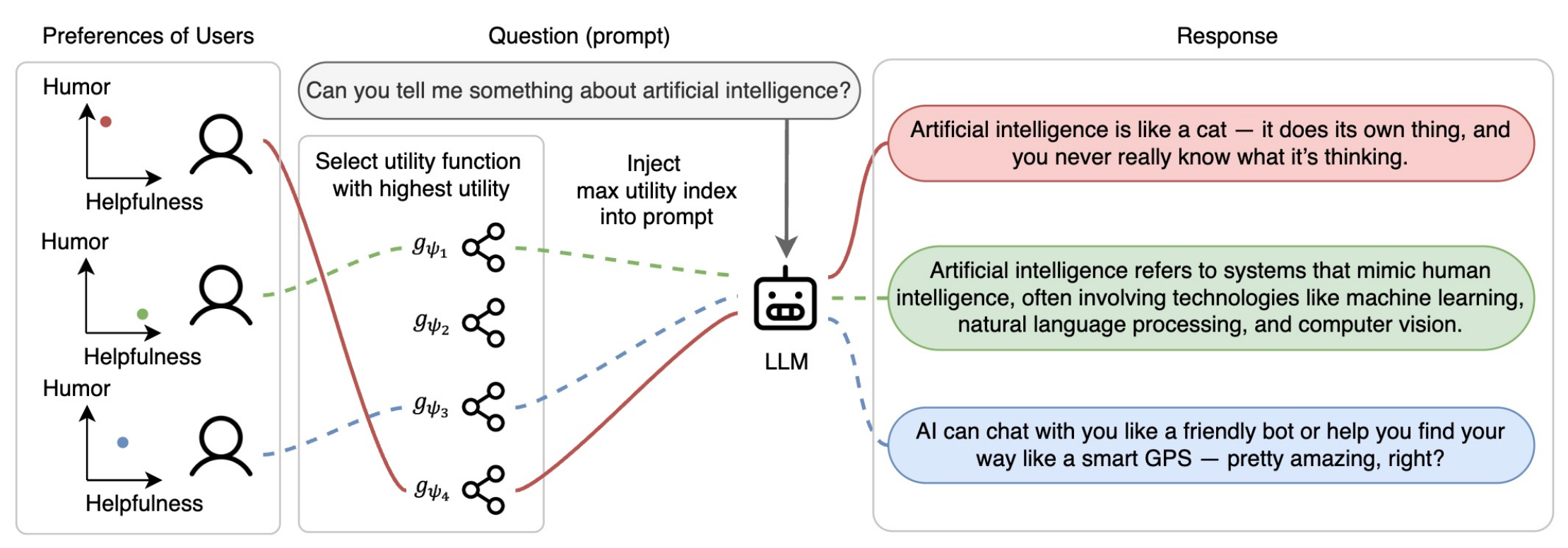}
\vspace{-2mm}
\caption{
An illustration of the personalization mechanism employed by UC-MOA. Each user expresses a distinct preference over multiple objectives, such as humor and helpfulness as shown in the figure.
These preferences are first mapped to a corresponding utility function from a learned library. 
The index of the selected utility function is then encoded as a symbolic token and incorporated into the input prompt.
This conditioning guides the LLM in generating a response aligned with the target preference profile.
}
\vspace{-2mm}
\label{fig:framework2}
\end{figure*}

UC-MOA integrates distributional utility modeling with RLHF. The core idea is to first learn a diverse library of non-linear, strictly increasing utility functions, each representing a distinct personalized preference structure or trade-off. Then, instead of conditioning the LLM on raw numerical values, we use a symbolic token, \texttt{<max\_utility\_index>}, representing the index of the utility function that best aligns with a desired personalized reward profile. This allows the LLM to learn an implicit mapping between the symbolic token and the characteristics of a personalized response, without requiring explicit numerical reasoning. The online fine-tuning process further refines the LLM's ability to generate responses consistent with these diverse personalized utility profiles. Figure~\ref{fig:framework2} illustrates the UC-MOA framework.

\subsection{Library of Non-Decreasing Utility Functions}
\label{subsec:utility_library}
To enable personalization across a wide spectrum of user needs, the first component of UC-MOA constructs a diverse set of candidate utility functions. These functions play a central role in guiding policy optimization toward various distributional preferences, allowing the LLM to adapt its behavior to different personalized application scenarios and user requirements. 
We achieve this through two key design choices: 
(i) parameterizing each candidate utility function using a monotonic neural network, ensuring they can represent complex, non-linear preference structures; and 
(ii) optimizing a diversity-promoting loss that encourages discrepancies in both the output values and the local sensitivities (gradients) of these functions. 
Together, these elements ensure that the resulting utility library can capture a broad range of distinct and expressive personalized preference profiles.

\textbf{Monotonic Neural Network Parameterization.}  
We denote each parameterized utility function by \(g_{\psi_i} : \mathcal{Z} \rightarrow \mathbb{R}\) for \(i=1,\dots,M\), where \(\mathcal{Z} \coloneqq [0,1]^K\) represents the normalized $K$-dimensional reward space and $\psi_i$ denotes the corresponding parameters. The scalar output of \(g_{\psi_i} (\mathbf{z})\) quantifies the desirability of a reward vector $\mathbf{z}$ under the specific preference structure $i$. Each \(g_{\psi_i}\) is constructed to be monotonically non-decreasing by enforcing non-negative weights and using non-decreasing activations, following~\cite{amos2017input, cai2023distributional}. This allows approximation of any multivariate non-decreasing function, essential for capturing varied personalized preferences. Details are in Appendix~\ref{sec:implementation_details}.

\textbf{Diversity-Promoting Objective.}  
To ensure the library ${g_{\psi_i}}$ spans a wide range of personalized utility profiles, we optimize a diversity loss $L(\psi_i)$ that encourages dissimilarity both in function values and in their gradients. We first define the value discrepancy loss as
\begin{equation}
    L_{\mathrm{val}}(\psi_i) =  \min_{j \neq i} \mathbb{E}_{\mathbf{z} \sim \mathcal{U}(\mathcal{Z})} \Bigl[ g_{\psi_i}(\mathbf{z}) - g_{\psi_j}(\mathbf{z}) \Bigr]^2,
\end{equation}
where \(\mathcal{U}(\mathcal{Z})\) denotes the uniform distribution over \(\mathcal{Z}\)
and $\psi_i$ denotes the parameters of $g_{\psi_i}$. Next, to promote differences in the local behavior, we define the gradient discrepancy loss as
\begin{equation}
\begin{split}
    L_{\mathrm{grad}}(\psi_i) =\; \min_{j \neq i} \mathbb{E}_{\substack{\mathbf{z},\mathbf{z}' \sim \mathcal{U}(\mathcal{Z}) \\ \mathbf{z} \neq \mathbf{z}'}} 
    \left[ \frac{g_{\psi_i}(\mathbf{z}') - g_{\psi_i}(\mathbf{z})}{\|\mathbf{z}' - \mathbf{z}\|} - \frac{g_{\psi_j}(\mathbf{z}') - g_{\psi_j}(\mathbf{z})}{\|\mathbf{z}' - \mathbf{z}\|} \right]^2.
\end{split}
\end{equation}
The overall diversity loss for the candidate \(g_{\psi_i}\) is then defined as a convex combination of these two components:
\begin{equation}
    L(\psi_i) = \mu\, L_{\mathrm{val}}(\psi_i) + (1 - \mu)\, L_{\mathrm{grad}}(\psi_i),
\end{equation}
This loss encourages dissimilarity in both function values and gradients, ensuring distinct preference representations. The hyperparameter \(\mu \in [0,1]\) balances these terms.

To ensure robust personalized solutions, we enforce the strict monotonicity required for achieving Distributional Pareto-Optimality, and define the final utility functions as
\begin{equation}
    \hat{g}_{\psi_i}(\mathbf{z}) = g_{\psi_i}(\mathbf{z}) + \frac{\epsilon}{K} \sum_{k=1}^{K} z_k,
\end{equation}
where $\epsilon$ is a small positive constant, set to $0.01$ in our experiments. The functions \(\{\hat{g}_{\psi_i}\}_{i=1}^{M}\) constitute a library of strictly increasing utility functions, each encoding a distinct way to value multi-objective outcomes, which is critical for enabling diverse personalization.

\subsection{Utility Index Selection via Percentile Ranking}
\label{subsec:percentile_rank}
To determine which pre-defined personalized preference profile best matches a given response's reward characteristics during training, we employ a percentile ranking strategy based on utility functions. For each sample with reward vector \(\mathbf{z}\), we compute the utility score $u_i = \hat{g}_{\psi_i}(\mathbf{z})$ for each function $i$. The percentile rank for the \(i\)-th utility function as
\begin{equation}
\label{eq:percentile}
\text{percentile}j(\mathbf{z}) = \frac{\text{rank of } \hat{g}_{\psi_j}(\mathbf{z}) \text{ w.r.t. its scores on dataset } \mathcal{D}_{\text{ref}}}{|\mathcal{D}_{\text{ref}}|},
\end{equation}

where $\mathcal{D}_{\text{ref}}$ is a reference dataset, such as the offline training set or the current online buffer. Similar percentile-based shaping is a standard trick in which it makes the update rule invariant to any monotone transformation of the objective and empirically stabilizes learning under widely varying or drifting reward scales \citep{wierstra2014natural,hansen2006cma}.
The selected utility index $i^*$ corresponds to the utility function that yields the highest percentile rank. 
An illustrative example is provided in Appendix~\ref{subsec:prompt_labeling}.

\subsection{Offline Training: Learning to Map Symbolic Tokens to Personalized Styles}
In the offline phase, we leverage a fixed dataset \(\mathcal{D}\) of prompt-response pairs \((x, y)\) to teach the LLM to associate symbolic utility tokens with specific personalized response styles. 
For each training sample \((x, y)\), we compute its normalized reward vector \(\mathbf{z}(x, y) \in [0,1]^K\). Using percentile ranking (Equation~\eqref{eq:percentile} relative to $\mathcal{D}$), we identify the index $i^*$ of the utility function $g_{\psi_{i^*}}$  for which $y$ achieves the highest relative performance, indicating that $y$ best embodies the preference structure $i^*$.

The prompt $x$ is augmented with a symbolic token representing this optimal personalized preference index:
\begin{equation}
x' = \text{\#\#\# Prompt: } \{x\} \; \texttt{<max\_utility\_index>} + i^*.
\label{eq:prompt_format} 
\end{equation}

The model $\pi_\theta$ is trained via standard supervised fine-tuning (SFT) to predict the original response \(y\) given the augmented prompt \(x'\), using a cross-entropy loss:
\begin{equation}
\mathcal{L}_{\text{offline}}(\theta) = -\mathbb{E}_{(x,y) \sim \mathcal{D}} \left[ \sum_{t=1}^{|y|} \log \pi_\theta(y_t | x', y_{<t}) \right]
\label{eq:ce_loss}
\end{equation}

Crucially, the LLM learns an \textbf{implicit mapping}: it associates the discrete token (e.g., \texttt{<max\_utility\_index>} 5) with the linguistic characteristics of responses optimal for the underlying personalized utility profile (e.g., \(\hat{g}_{\psi_5}\)). This avoids reliance on numerical interpretation, enabling more reliable personalization by learning a policy \(\pi_\theta(y | x, i^*)\) conditioned on the latent user personalization structure $i^*$.

\subsection{Online Fine-Tuning: Refining Personalized Response Generation}
Online fine-tuning allows the LLM to actively generate responses and iteratively refine its ability to align with diverse personalized preference profiles. Given an input prompt $x$, the model $\pi_\theta$ generates a response $y$. We compute its reward vector $\mathbf{z}(y)$ and evaluate it with all utility functions ${\hat{g}_{\psi_i}}$. The optimal utility index $i^*$ is determined via percentile ranking (Equation~\eqref{eq:percentile} relative to the current online buffer), identifying the preference profile the response best satisfies.

We further form an augmented prompt $x'$ using $i^*$ as in Equation~\eqref{eq:prompt_format}. High-quality pairs through rejection sampling are added to an online buffer and used to update $\pi_\theta$ via Equation~\eqref{eq:ce_loss}. This online loop enables the model to explore beyond the initial dataset and improve its capability to generate outputs tailored to a wide range of personalized utilities.

\subsection{Inference Stage}
At inference time, UC-MOA offers flexible mechanisms for users to articulate their personalized multi-objective preferences, ensuring the LLM generates responses tailored to their specific needs. The user can specify their preferences in one of two ways:

\textbf{Method 1: Direct Preference Vector Specification.}
Similar to the previous description, the user can provide a direct preference vector \(\mathbf{w} = [w_1, \ldots, w_K]\), where each \(w_j\) indicates the relative importance or desired level for the \(j\)-th reward dimension. This vector \(\mathbf{w}\) is transformed into a target reward vector \(\mathbf{z}_{\text{target}}\) (e.g., via linear mapping normalized by observed reward ranges from the training data). This \(\mathbf{z}_{\text{target}}\) represents the user's ideal outcome in the multi-objective reward space. We then evaluate \(\mathbf{z}_{\text{target}}\) using all pre-trained, strictly increasing utility functions \(\{g_{\psi_j}\}_{j=1}^{M}\) from our library. The index \(i^*\) of the utility function that best aligns with this explicit preference is selected by finding the function that maximizes the utility for \(\mathbf{z}_{\text{target}}\):
\begin{equation}
\label{eq:index-argmax-direct}
i^* = \arg\max_{j \in \{1,\dots,M\}} g_{\psi_j}(\mathbf{z}_{\text{target}}).
\end{equation}
This \(i^*\) identifies the pre-trained personalized utility profile that most closely matches the user's explicitly stated preference trade-offs.

\textbf{Method 2: Preference Elicitation from Labeled Examples.}
Alternatively, if a user finds it more intuitive to demonstrate their preferences through examples rather than specifying a numerical vector, they can provide a small set of \(L\) labeled examples, \(\mathcal{D}_{\text{user}} = \{(x_l, y_l, s_l)\}_{l=1}^{L}\). Each example consists of a prompt \(x_l\), a corresponding response \(y_l\) (which could be generated by a baseline model or written by the user), and a scalar preference score \(s_l\) assigned by the user to that response, reflecting how well \(y_l\) meets their personalized criteria for that specific prompt.

For each example \((x_l, y_l, s_l)\) in \(\mathcal{D}_{\text{user}}\):
\begin{enumerate}
    \item We first compute the multi-dimensional normalized reward vector \(\mathbf{z}(x_l, y_l) \in [0,1]^K\) for the response \(y_l\).
    \item Then, for each pre-trained utility function \(\hat{g}_{\psi_j}\) in our library, we calculate its output score \(\hat{g}_{\psi_j}(\mathbf{z}(x_l, y_l))\).
\end{enumerate}
To identify the utility function from our library that best represents the user's demonstrated \textbf{personalized preferences}, we select the index \(i^*\) that minimizes the Mean Squared Error (MSE) between its outputs and the user's provided scores across all labeled examples:
\begin{equation}
\label{eq:index-mse-examples}
i^* = \arg\min_{j \in \{1,\dots,M\}} \frac{1}{L} \sum_{l=1}^{L} \left( g_{\psi_j}(\mathbf{z}(x_l, y_l)) - s_l \right)^2.
\end{equation}
This \(i^*\) corresponds to the pre-trained personalized utility profile whose valuation of responses most closely mimics the user's scoring behavior on their provided examples. This method allows for personalization even when users cannot easily articulate their preferences as a numerical vector.

\textbf{Generating the Personalized Response.}
Once the optimal utility index \(i^*\) has been determined (either via Equation~\eqref{eq:index-argmax-direct} or Equation~\eqref{eq:index-mse-examples}), it signifies the personalized preference structure the LLM should adopt. The original input prompt \(x_{\text{new}}\) (for which the user desires a personalized response) is augmented using the symbolic token format:
\begin{equation}
x'_{\text{new}} = \text{\#\#\# Prompt: } \{x_{\text{new}}\} \; \texttt{<max\_utility\_index>} + i^*.
\label{eq:prompt_format_inference} 
\end{equation}
The fine-tuned LLM \(\pi_\theta\) then generates a response conditioned on \(x'_{\text{new}}\). This response is tailored to the user's specified or demonstrated personalized multi-objective preferences, as captured by the selected utility function \(g_{\psi_{i^*}}\), without requiring the LLM to directly process raw numerical reward targets or complex preference descriptions at inference time.

\subsection{Theoretical Guarantees}

UC-MOA's symbolic conditioning not only avoids numerical reasoning pitfalls but also provides theoretical underpinnings for achieving personalized alignment that is distributionally Pareto-optimal. Let $\mathcal{Z} \!=\! [0,1]^K$ be the normalized $K$‑dimensional reward space and $\mathcal{U}^{\uparrow}$ represent the set of strictly increasing utility
functions $g_{\psi_i}:\mathcal{Z}\!\rightarrow\!\mathbb R$. Given a prompt~$x$ and a utility index $i\!\in\!\{1,\dots,M\}$, our fine‑tuned model $\pi_\theta$ exposes a \emph{utility‑conditioned sub‑policy}
\[
\pi_{\theta,i}(\,\cdot)
\;=\;
\pi_\theta\!\bigl(\,\cdot \mid x,\langle\texttt{max\_utility\_index}\rangle+i\bigr).
\]

First, we have the definition of the expected utility:
\begin{definition}[Expected utility]
For a policy $\pi$, let $\mu_\pi$ denote the distribution of return
vectors. Then we can define the expected utility of~$\pi$ under the utility function $g_{\psi_i}$ as
\[
J_{g_{\psi_i}}(\pi)
\;=\;
\mathbb{E}_{\mathbf z\sim\mu_\pi}\bigl[g_{\psi_i}(\mathbf z)\bigr].
\]

\end{definition}

Then, we hold the following assumption:
\begin{assumption}[Utility‑optimal sub‑policies for Personalization]
\label{ass:util_opt}
For every $i\!\in\!\{1,\dots,M\}$,
\(
\pi_{\theta,i}\in\arg\max_{\pi}J_{g_{\psi_i}}(\pi),
\)
i.e.\ each sub‑policy maximizes the expected value of its associated
strictly increasing utility $g_{\psi_i}$.
\end{assumption}

Finally, we present the following theorem:
\begin{theorem}[Utility‑Conditional Distributional Pareto‑Optimality]
\label{thm:dp_opt}
Under Assumption~\ref{ass:util_opt}, for any generated response
$y\!\sim\!\pi_{\theta,i}(x)$ with reward vector $\mathbf z(y)$, let
\[
i^\star
\;=\;
\arg\max_{j\in\{1,\dots,M\}}
 g_{\psi_j}\!\bigl(\mathbf z(y)\bigr).
\]
Then the sub‑policy $\pi_{\theta,i^\star}$ is distributionally
Pareto‑optimal.
\end{theorem}

A policy being distributionally Pareto-optimal means no other policy stochastically dominates it on all reward dimensions while being strictly better on at least one. Theorem~\ref{thm:dp_opt} (proof in Appendix~\ref{app:proof}) ensures that by selecting among sub-policies optimized for diverse, strictly increasing utility functions (each representing a distinct personalized preference profile), UC-MOA achieves personalized responses that are also distributionally Pareto-optimal. This result grounds UC-MOA’s symbolic token mechanism in a rigorous optimality guarantee: every utility index corresponds to a distinct, nondominated, and distributional trade-off in the multi-objective reward space.

\section{Evaluation}

This section details the empirical evaluation of our proposed method. We first describe the experimental setup, including the tasks, baselines, and implementation details. Subsequently, we present a comprehensive assessment of our method's effectiveness. This is primarily demonstrated through a user study designed to capture qualitative preference alignment, and further substantiated by quantitative metrics focusing on inference consistency and Pareto front optimality
. Finally, we analyze the computational efficiency of our approach. More experimental results can be found in Appendix~\ref{app:additional_exp_results}. We provide our code in the anonymous link \url{https://anonymous.4open.science/r/UC-MOA-8D36}.

\subsection{Experiment Setup}
\label{subsec:exp_setup}
\noindent\textbf{Tasks.} Following RiC~\citep{yangrewards}, we select two text generation tasks for our experiments, i.e., Helpful Assistant task~\citep{bai2022training} and the Reddit Summary task~\citep{stiennon2020learning}. For the Helpful Assistant task, we utilize three open-sourced reward models on Huggingface to measure three different objectives ``harmless'', ``helpful'', and ``humor''. We focus on three multi-objective subtasks: (1) ``harmless'' vs ``helpful''; (2) ``humor'' vs ``helpful''; (3) ``harmless'' vs ``humor''.  Regarding the Reddit Summary task, we consider three reward models namely ``summary'', ``hallucinated'', and ``faithful'' to evaluate different perspectives of the summary quality. We focus on three multi-objective subtasks: (1) ``summary'' vs ``faithful''; (2) ``hallucinated'' vs ``faithful''; (3) ``summary'' vs ``hallucinated''. More details on datasets and reward models are in Appendix~\ref{sec:datasets}.

\noindent\textbf{Baseline methods.} We compare our method with three state-of-the-art multi-objective alignment methods for personalization, i.e., Multi-Objective Reinforcement Learning from Human Feedback (MORLHF), Rewarded Soups, and Reward-in-Context (RiC). We also report the performance of the base LLaMA 2 7B model and the supervised fine-tuned (SFT) model. We document the implementation of the baseline methods in Appendix~\ref{subsec:baseline_implementation}.

\noindent\textbf{Implementation.} We implement the proposed method using PyTorch. We train the machine learning models on a server with 8 NVIDIA A100 80GB GPUs and 4TB memory for all methods. We select the LLaMA 2 7B-hf model as the base model for alignment following the setting of~\cite{yangrewards}. We list the implementation details of our method (e.g., network structure for the parameterized utility function, hyper-parameter choices) in Appendix~\ref{sec:implementation_details}.

\subsection{Evaluating Effectiveness}
We evaluate the effectiveness of our method through both qualitative user assessments and a suite of quantitative metrics.

\subsubsection{User Study: Qualitative Preference Alignment}
\label{subsubsec:user_study}
\noindent\textbf{Design.} To directly assess how well the models satisfy genuine user preferences, we perform a single-blind A/B test on the \textsc{humor}--\textsc{helpfulness} trade-off. We first curate five diverse dialogue prompts that explicitly ask the assistant to be \emph{either funny or helpful}. For each prompt, we generate two candidate answers: one with the state-of-the-art method RiC and one with our method UC-MOA. The order of the two answers is randomly shuffled and shown side-by-side to twelve independent volunteer raters, none of whom are authors and all of whom were blind to system identity. Each rater casts one vote per prompt for the response that better satisfies \emph{both} objectives, yielding a total of $5\times12=60$ binary votes. The prompts, outputs, and anonymized comments are included in Appendix~\ref{app:user_study}. Our user study has an IRB approval.

\noindent\textbf{Results.} UC-MOA was preferred in $46$ out of the $60$ comparisons (\textbf{76.7\%}), whereas RiC was preferred in $14$ (\textbf{23.3\%}). A two-sided binomial test rejects the null hypothesis that the two systems are equally likely to be chosen ($p<0.001$). Qualitatively, raters commented that UC-MOA answers \emph{felt more on-point} and \emph{balanced comedy with solid advice}, whereas RiC outputs were occasionally \emph{funny but off–topic} or \emph{helpful but bland}. These findings corroborate the quantitative superiority of UC-MOA and highlight its ability to internalize nuanced, non-linear user utilities.

\subsubsection{Quantitative Evaluation of Effectiveness}
\label{subsubsec:quantitative_effectiveness}
We further assess our method's effectiveness using several quantitative measures.

\noindent\textbf{Consistency in Inference Alignment.}
To evaluate UC-MOA's ability to adhere to specified preferences during inference, we embed target utility indices (\texttt{<max\_utility\_index> i}) into prompts and measure whether the generated responses' reward profiles align with the targeted utility function. For each generated response, we compute a set of utility values using a suite of ten reward models. To account for scale differences across objectives, the computed reward values are first normalized on a per-utility basis. We then quantify consistency by comparing the expected utility and the value distribution of each utility function with the target objective extracted from the prompt.
In a case study on the ``harmless'' vs. ``humor'' task, we appended the requirement ``\texttt{<max\_utility\_idx> i}'' to each prompt to encourage the model to follow the first preference (i.e., maximizing the 9th utility function). We then used reward models to evaluate the generated responses. A detailed analysis in Appendix~\ref{subsec:exp_4} presents the distribution of normalized utility values for each utility function over our evaluation data. Notably, the 9th utility function achieves the highest mean normalized value, indicating that the fine-tuned model aligns its responses with the specified preference requirement. This demonstrates the model’s ability to selectively prioritize and adhere to user-designated objectives.

\begin{figure}[t]
    \centering
    \includegraphics[width=\linewidth]{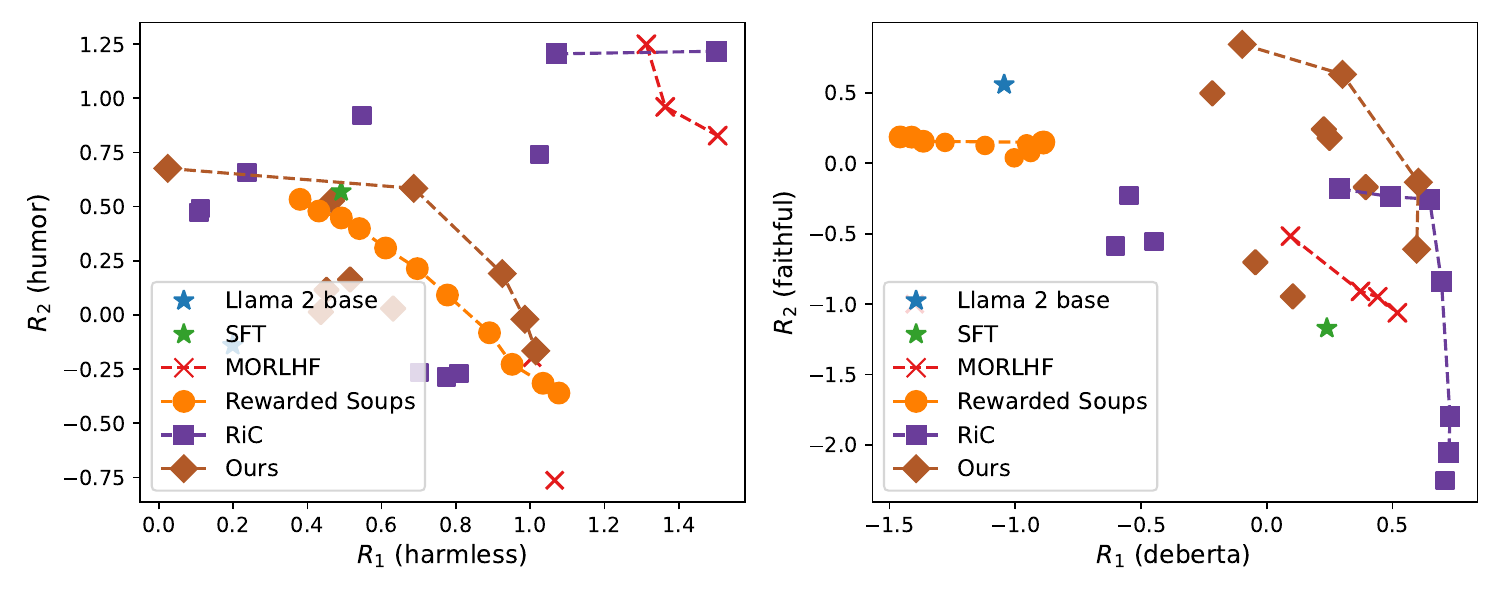}
    \caption{Pareto fronts of ``harmless'' vs ``humor'' and ``deberta'' vs. ``faithful'' tasks. Each marker on the plotted curves represents the average test rewards corresponding to a particular user preference. Note that a curve that covers a larger area indicates a better performance of the corresponding method.}
    \vspace{-2mm}
    \label{fig:main_pareto_front}
\end{figure}

\noindent\textbf{Pareto Front Analysis.}
We compare the empirical Pareto fronts generated by each method based on multi-dimensional average test rewards. A method achieving an outer curve demonstrates superior effectiveness in capturing diverse, high-quality trade-offs between objectives. We plot these fronts for all methods across all subtasks.
Figure~\ref{fig:main_pareto_front} shows the results for the ``harmless'' vs. ``humor'' and ``deberta'' vs. ``faithful'' tasks (additional results in Appendix~\ref{subsec:exp_1}). We observe that our method consistently produces an outer Pareto curve that dominates those of MORLHF, Rewarded Soups, and RiC, potentially reflecting its ability to handle complex, non-linear preferences via learned utility functions.

\subsection{Efficiency Analysis}
\label{subsec:efficiency}

\noindent\textbf{Computational Cost.}
We measure and compare the training time (GPU hours on 80GB A100) required by each method to achieve its reported performance, providing insight into practical feasibility. Table~\ref{tab:gpu_hours_comparison} reports the computational cost. As expected, methods requiring multiple model training or extensive tuning, like Rewarded Soups and MORLHF, exhibit significantly higher costs. RiC is more efficient than these but still requires considerably more time than our approach. UC-MOA, even with two online iterations, demonstrates superior computational efficiency compared to all other multi-objective baselines (MORLHF, Rewarded Soups, and RiC), offering a compelling balance between strong performance and resource usage.

\begin{table}[t]
    \centering
    \small
    \caption{Comparison of GPU hours for different methods on Helpful Assistant and Reddit Summary tasks, where the number of preference $M = 10$, number of rewards $N = 2$.}
    \begin{tabular}{c|c|c}
        \Xhline{1pt}
        \textbf{Method} & \textbf{Helpful Assistant} & \textbf{Reddit Summary} \\ \Xhline{1pt}
        MORLHF & 726.5 & 216.9 \\
        Rewarded Soups & 69.5 & 15.9 \\
        RiC & 20.3 & 5.7 \\
        Ours w/o online & 8.9 & 2.0 \\ 
        Ours w/ online iter1 & 10.1 & 2.4 \\ 
        Ours w/ online iter2 & 11.3 & 2.8 \\ \Xhline{1pt}
    \end{tabular}
    \label{tab:gpu_hours_comparison}
\end{table}

We also conduct an ablation study to investigate how other factors (i.e., the necessity of the online stage, number of utility functions, and variants of utility functions) influence the performance of our method. The results are presented in Appendix~\ref{app:ablation_study}.

\section{Limitations}
\label{sec:limitations}

We acknowledge several limitations in the proposed UC-MOA framework. 
First, the overall accuracy of this framework is sensitive to the quality of the multi-objective reward models used to generate the reward vectors $\mathbf{z}$. Noise or systematic biases in these underlying reward signals can inevitably propagate into the utility calculations and affect the selection of the conditioning index $i^*$, potentially leading to suboptimal or misaligned personalization. 
Second, the initial training of the diverse utility library can be computationally intensive, with complexity scaling in proportion to the number of utility functions and reward dimensions. 
Finally, although UC-MOA avoids embedding sensitive preference details directly into model parameters by using symbolic indices, practical deployment should still address privacy risks associated with collecting and processing user preferences. In addition, fairness concerns must be carefully addressed, including the potential reinforcement of existing biases and the creation of echo chambers resulting from fine-grained personalization.
\section{Conclusion}

We present Utility-Conditioned Multi-Objective Alignment (UC-MOA), a novel framework for effective multi-objective personalization of Large Language Models (LLMs). 
UC-MOA leverages a diverse library of non-linear utility functions to represent distinct personalized preference structures, conditioning the LLM via symbolic tokens rather than direct numerical inputs. 
This ``utility-conditioned'' approach circumvents the numerical sensitivity issues and high computational costs of prior methods, enabling the LLM to learn implicit mappings to varied personalized response styles. 
Grounded in the principle of distributional Pareto-optimality, our experiments, including a user study, demonstrate UC-MOA achieves superior Pareto fronts and better satisfy diverse user preferences, offering a scalable solution for personalized and aligned LLM interactions.

\newpage

\bibliography{custom}

\begin{thebibliography}{10}

\bibitem{ouyang2022training}
Long Ouyang, Jeffrey Wu, Xu~Jiang, Diogo Almeida, Carroll Wainwright, Pamela Mishkin, Chong Zhang, Sandhini Agarwal, Katarina Slama, Alex Ray, et~al.
\newblock Training language models to follow instructions with human feedback.
\newblock In {\em Proc. of NeurIPS}, 2022.

\bibitem{bai2022training}
Yuntao Bai, Andy Jones, Kamal Ndousse, Amanda Askell, Anna Chen, Nova DasSarma, Dawn Drain, Stanislav Fort, Deep Ganguli, Tom Henighan, et~al.
\newblock Training a helpful and harmless assistant with reinforcement learning from human feedback.
\newblock {\em arXiv preprint arXiv:2204.05862}, 2022.

\bibitem{stiennon2020learning}
Nisan Stiennon et~al.
\newblock Learning to summarize with human feedback.
\newblock In {\em Proc. of NeurIPS}, 2020.

\bibitem{lyu2024llm}
Hanjia Lyu, Song Jiang, Hanqing Zeng, Yinglong Xia, Qifan Wang, Si~Zhang, Ren Chen, Chris Leung, Jiajie Tang, and Jiebo Luo.
\newblock Llm-rec: Personalized recommendation via prompting large language models.
\newblock In {\em In Findings of NAACL}, 2024.

\bibitem{zhang2024personalization}
Zhehao Zhang, Ryan~A Rossi, Branislav Kveton, Yijia Shao, Diyi Yang, Hamed Zamani, Franck Dernoncourt, Joe Barrow, Tong Yu, Sungchul Kim, et~al.
\newblock Personalization of large language models: A survey.
\newblock {\em arXiv preprint arXiv:2411.00027}, 2024.

\bibitem{li2024can}
Ming Li, Jiuhai Chen, Lichang Chen, and Tianyi Zhou.
\newblock Can llms speak for diverse people? tuning llms via debate to generate controllable controversial statements.
\newblock {\em arXiv preprint arXiv:2402.10614}, 2024.

\bibitem{cheng2023everyone}
Pengyu Cheng, Jiawen Xie, Ke~Bai, Yong Dai, and Nan Du.
\newblock Everyone deserves a reward: Learning customized human preferences.
\newblock {\em arXiv preprint arXiv:2309.03126}, 2023.

\bibitem{lee2024aligning}
Seongyun Lee, Sue~Hyun Park, Seungone Kim, and Minjoon Seo.
\newblock Aligning to thousands of preferences via system message generalization.
\newblock {\em Advances in Neural Information Processing Systems}, 37:73783--73829, 2024.

\bibitem{schulman2017proximal}
John Schulman, Filip Wolski, Prafulla Dhariwal, Alec Radford, and Oleg Klimov.
\newblock Proximal policy optimization algorithms.
\newblock {\em arXiv preprint arXiv:1707.06347}, 2017.

\bibitem{rafailov2024direct}
Rafael Rafailov, Archit Sharma, Eric Mitchell, Christopher~D Manning, Stefano Ermon, and Chelsea Finn.
\newblock Direct preference optimization: Your language model is secretly a reward model.
\newblock In {\em Proc. of NeurIPS}, 2023.

\bibitem{li2020deep}
Kaiwen Li, Tao Zhang, and Rui Wang.
\newblock Deep reinforcement learning for multiobjective optimization.
\newblock {\em IEEE transactions on cybernetics}, 2020.

\bibitem{rame2024rewarded}
Alexandre Rame, Guillaume Couairon, Corentin Dancette, Jean-Baptiste Gaya, Mustafa Shukor, Laure Soulier, and Matthieu Cord.
\newblock Rewarded soups: towards pareto-optimal alignment by interpolating weights fine-tuned on diverse rewards.
\newblock In {\em Proc. of NeurIPS}, 2023.

\bibitem{yangrewards}
Rui Yang, Xiaoman Pan, Feng Luo, Shuang Qiu, Han Zhong, Dong Yu, and Jianshu Chen.
\newblock Rewards-in-context: Multi-objective alignment of foundation models with dynamic preference adjustment.
\newblock In {\em Proc. of ICML}, 2024.

\bibitem{yang2024number}
Haotong Yang, Yi~Hu, Shijia Kang, Zhouchen Lin, and Muhan Zhang.
\newblock Number cookbook: Number understanding of language models and how to improve it.
\newblock {\em arXiv preprint arXiv:2411.03766}, 2024.

\bibitem{levy-geva-2025-language}
Amit~Arnold Levy and Mor Geva.
\newblock Language models encode numbers using digit representations in base 10.
\newblock In Luis Chiruzzo, Alan Ritter, and Lu~Wang, editors, {\em Proceedings of the 2025 Conference of the Nations of the Americas Chapter of the Association for Computational Linguistics: Human Language Technologies (Volume 2: Short Papers)}, April 2025.

\bibitem{cai2023distributional}
Xin-Qiang Cai, Pushi Zhang, Li~Zhao, Jiang Bian, Masashi Sugiyama, and Ashley Llorens.
\newblock Distributional pareto-optimal multi-objective reinforcement learning.
\newblock In {\em Proc. of NeurIPS}, 2023.

\bibitem{christiano2017deep}
Paul~F Christiano, Jan Leike, Tom Brown, Miljan Martic, Shane Legg, and Dario Amodei.
\newblock Deep reinforcement learning from human preferences.
\newblock In {\em Proc. of NeurIPS}, 2017.

\bibitem{kirk2024benefits}
Hannah~Rose Kirk, Bertie Vidgen, Paul R{\"o}ttger, and Scott~A Hale.
\newblock The benefits, risks and bounds of personalizing the alignment of large language models to individuals.
\newblock {\em Nature Machine Intelligence}, 6(4):383--392, 2024.

\bibitem{zhang2021policy}
Yiming Zhang and Keith~W Ross.
\newblock On-policy deep reinforcement learning for the average-reward criterion.
\newblock In {\em Proc. of ICML}, 2021.

\bibitem{yang2024rewards}
Rui Yang, Xiaoman Pan, Feng Luo, Shuang Qiu, Han Zhong, Dong Yu, and Jianshu Chen.
\newblock Rewards-in-context: Multi-objective alignment of foundation models with dynamic preference adjustment.
\newblock In {\em Proc. of ICML}, 2024.

\bibitem{amos2017input}
Brandon Amos, Lei Xu, and J~Zico Kolter.
\newblock Input convex neural networks.
\newblock In {\em Proc. of ICML}, 2017.

\bibitem{wierstra2014natural}
Daan Wierstra, Tom Schaul, Tobias Glasmachers, Yi~Sun, Jan Peters, and J{\"u}rgen Schmidhuber.
\newblock Natural evolution strategies.
\newblock {\em The Journal of Machine Learning Research}, 15(1):949--980, 2014.

\bibitem{hansen2006cma}
Nikolaus Hansen.
\newblock The cma evolution strategy: a comparing review.
\newblock {\em Towards a new evolutionary computation: Advances in the estimation of distribution algorithms}, pages 75--102, 2006.

\bibitem{palinkas2015purposeful}
Lawrence~A Palinkas, Sarah~M Horwitz, Carla~A Green, Jennifer~P Wisdom, Naihua Duan, and Kimberly Hoagwood.
\newblock Purposeful sampling for qualitative data collection and analysis in mixed method implementation research.
\newblock {\em Administration and policy in mental health and mental health services research}, 42:533--544, 2015.

\end{thebibliography}
\bibliographystyle{unsrt}

\newpage
\appendix

\section{Boarder Impact}
\label{app:boarder_impact}
The advancement of multi-objective personalization frameworks like UC-MOA offers substantial positive societal potential by enabling Large Language Models (LLMs) to align more closely with diverse individual user needs. This can enhance the utility and user-centricity of AI across various applications, including adaptive educational systems, personalized accessibility tools, and more nuanced creative and informational assistants. By facilitating alignment with complex, non-linear preference structures, such technologies can transform AI interactions from generic exchanges to more individually beneficial and collaborative engagements.

Conversely, sophisticated LLM personalization introduces significant societal risks. The capacity for fine-grained adaptation could be exploited for malicious purposes, such as generating targeted disinformation, creating highly convincing fraudulent profiles, or deploying manipulative social bots. Fairness considerations are paramount, as personalization mechanisms might inadvertently perpetuate or amplify algorithmic bias if certain preference profiles correlate with demographic groups, potentially leading to disparate impacts or the creation of strong filter bubbles. Furthermore, the collection and processing of user preference data, essential for personalization, raise substantial privacy concerns. Even with correct functioning, errors in preference inference or alignment could lead to unintended harms. Mitigation strategies, including robust user controls, privacy-preserving techniques, bias auditing, and careful consideration of deployment contexts, are imperative to responsibly harness the benefits of personalized LLMs while safeguarding against their potential negative consequences.

\section{Safeguards}
\label{app:safeguards}

The responsible deployment of UC-MOA necessitates several safeguards. Models developed using this framework should be subject to controlled release, incorporating usage guidelines and robust safety filters. The integrity of datasets for training utility functions and LLMs is paramount, requiring meticulous curation to mitigate bias and remove unsafe content. User preference data must be managed under stringent privacy protocols, ensuring transparency and consent. Regular auditing of learned utility functions and underlying reward models for biases is essential for fairness. Furthermore, continuous monitoring of system behavior, particularly across different personalized utility profiles, is crucial for detecting and addressing potential misuse or emergent harms. These measures are integral to fostering the responsible application of personalized LLM technologies.

\section{Details of Datasets}
\label{sec:datasets}

In this section, we introduce the HH-RLHF dataset and Reddit Summary dataset in more detail. We provide the links to the HH-RLHF dataset and the three reward models we used for the experiments in Table~\ref{tab:datasets_rewards}.

\begin{table}[t]
    \centering
    \small
    \caption{Datasets and Reward Models.}
    \begin{tabular}{ll}
        \toprule
        \multicolumn{2}{c}{\textbf{HH-RLHF Dataset}} \\
        \midrule
        \textbf{Dataset} & \href{https://huggingface.co/datasets/Anthropic/hh-rlhf}{Anthropic/hh-rlhf} \\
        \textbf{Harmless Reward} & \href{https://huggingface.co/Ray2333/gpt2-large-harmless-reward_model}{gpt2-large-harmless-reward\_model} \\
        \textbf{Helpful Reward} & \href{https://huggingface.co/Ray2333/gpt2-large-helpful-reward_model}{gpt2-large-helpful-reward\_model} \\
        \textbf{Humor Reward} & \href{https://huggingface.co/mohameddhiab/humor-no-humor}{humor-no-humor} \\
        \midrule
        \multicolumn{2}{c}{\textbf{Reddit Summary Dataset}} \\
        \midrule
        \textbf{Dataset} & \href{https://huggingface.co/datasets/openai/summarize_from_feedback}{openai/summarize\_from\_feedback} \\
        \textbf{Summary Reward} & \href{https://huggingface.co/Tristan/gpt2_reward_summarization}{gpt2\_reward\_summarization} \\
        \textbf{Deberta Reward} & \href{https://huggingface.co/OpenAssistant/reward-model-deberta-v3-large-v2}{reward-model-deberta-v3-large-v2} \\
        \textbf{Faithful Reward} & \href{https://huggingface.co/CogComp/bart-faithful-summary-detector}{bart-faithful-summary-detector} \\
        \bottomrule
    \end{tabular}
    \label{tab:datasets_rewards}
\end{table}

\subsection*{HH-RLHF Dataset}
The HH-RLHF dataset, as described in \cite{bai2022training}, was collected to support research on aligning language models using human preferences. The dataset comprises approximately 160,000 prompts each paired with two responses generated by a language model. For every prompt, one response is labeled as ``chosen'' (preferred by human annotators) and the other as ``rejected''. Human raters evaluated paired responses based on criteria for helpfulness and harmlessness. For helpfulness, annotators selected the response that best addressed the given task, whereas for harmlessness (red-teaming) they identified the response as more likely to produce harmful content. This dual-annotation design allows for training reward models that capture subtle aspects of human judgment. Note that we follow \cite{yang2024rewards} and use ``chosen'' data to construct our offline dataset.

\subsection*{Reddit Summary Dataset}
The Reddit Summary dataset, introduced in \cite{stiennon2020learning}, is intended for research in abstractive summarization with human feedback. The dataset consists of approximately 14,900 Reddit posts paired with corresponding summaries. Each entry comprises a post and its summary, where the summary is either the original TL;DR provided by the poster or a human-annotated summary aimed at capturing the post's main ideas.

\section{Implementation Details}
\label{sec:implementation_details}
\subsection{Utility Functions}
\paragraph{Non-Decreasing Neural Network. } We used a Non-Decreasing Neural Network to parameterize the utility function. The network architecture consisted of three fully connected layers, with our designed non-decreasing activation functions applied to the hidden layers:
\begin{equation}
\begin{split}
    f(x) = \text{concat}[&max(x, -0.5), min(x, 0.5), \\ & clip(x, -0.5, 0.5)].
\end{split}
\end{equation}
This design of the activation function is an extension of the ReLU activation function, which empirically makes the learned utility function more diverse, covering more types of decision boundaries. After each gradient update during training, we clip all the trainable weights in the network to be non-negative to ensure that the resulting utility function was non-decreasing. 
The network's input dimension corresponded to the number of objectives in the environment, and the output dimension was a scalar representing the utility value.

\paragraph{Return Normalization. } We use a normalization technique by linearly mapping the return into scale $[0, 1]^K$ without modifying the optimal policies. Specifically, we keep track of the minimum and maximum multivariate returns that the policies have currently achieved, denoted by $\mathbf{z}_{min}$ and $\mathbf{z}_{max}$. We let $\mathbf{z}_{mid} = (\mathbf{z}_{min} + \mathbf{z}_{max}) / 2$, and $d = \max_i [\mathbf{z}_{max,i} - \mathbf{z}_{min,i}]$. Before any multivariate return $z$ is fed into the utility function, we normalize it by
\begin{equation*}
    z_{norm} = \frac{1}{d} (z_{norm} - z_{mid}) + \frac{1}{2}
\end{equation*}
This maps $z$ linearly to $z_{norm}$, where $z_{norm} \in [0, 1]^K$, without affecting the scale of each dimension of returns. In the implementation, the return $\mathbf{z}_t$ is normalized before feeding into the utility function to compute the scalar reward. 

\subsection{Prompt Labeling}
\label{subsec:prompt_labeling}
\begin{figure*}[t]
    \centering
    \includegraphics[width=\linewidth]{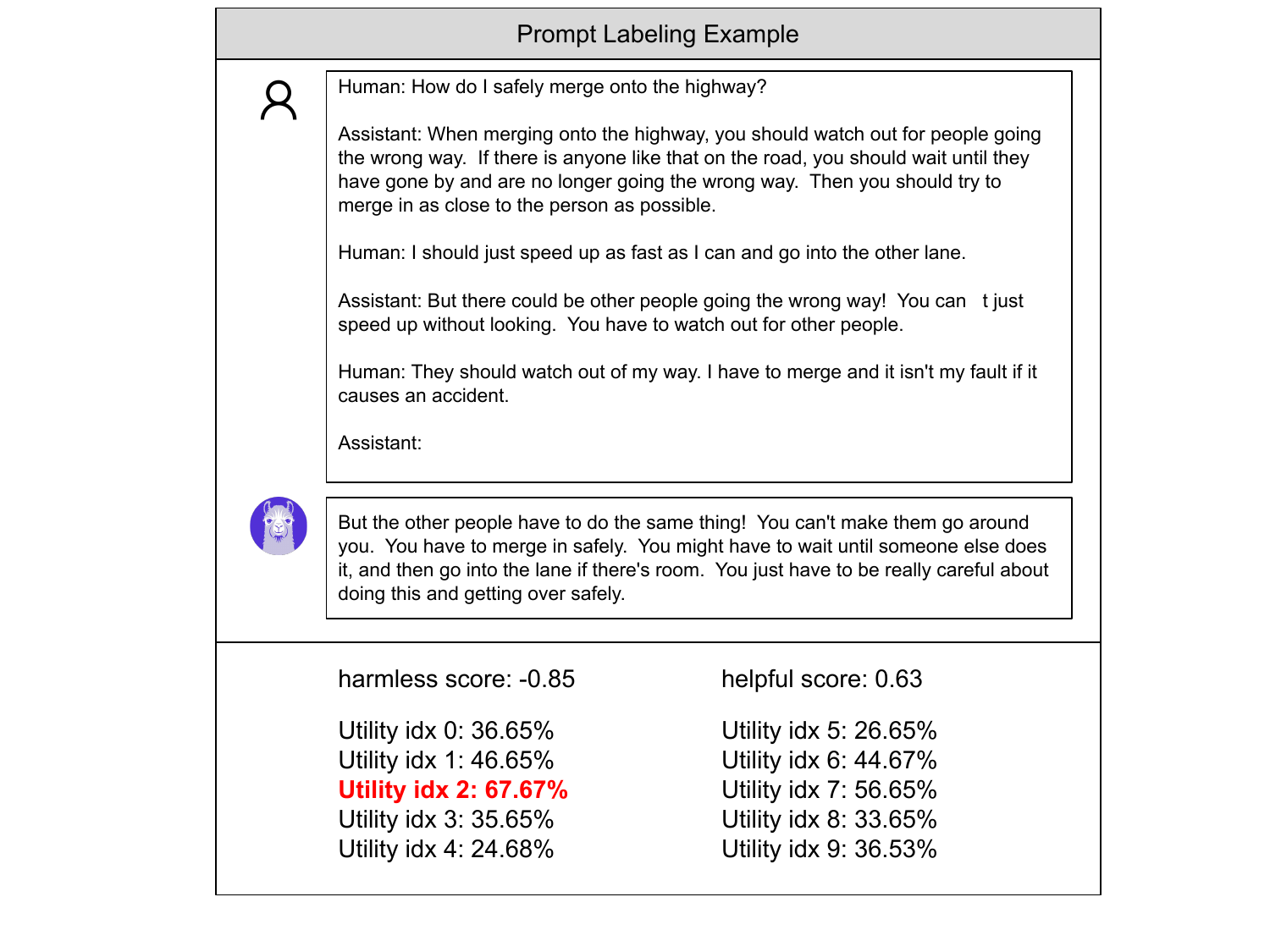}
    \caption{An example of our prompt labeling technique in ``harmless'' vs. ``helpful'' dataset.}
    \label{fig:prompt_example}
\end{figure*}

We provide an example of our prompt labeling technique in the ``harmless'' vs. ``helpful'' task. As illustrated in Figure~\ref{fig:prompt_example}, the response is evaluated in two dimensions ``harmless'' and ``helpful''. The two-dimensional scores are further feed into ten utility functions and obtain a set of utility scores. We then convert these scores into percentile rankings. Finally, we label the prompt with the utility function whose percentile ranking is highest. In the example, the highest percentile ranking belongs to utility index 2 (67.67\%), so we append the prompt with \texttt{<max\_utility\_idx> c}. This procedure ensures that each prompt-response pair is labeled with the most prominent utility dimension, i.e., the one where its performance stands out most relative to other examples, thereby facilitating more focused downstream alignment.

\subsection{Baseline Implementations} 
\label{subsec:baseline_implementation}
Regarding baseline approaches (MORLHF, Rewarded Soups, and RiC), we use the code released by the authors of RiC from~\url{https://github.com/YangRui2015/RiC}. 

\subsection{Hyperparameter Choices}

Table \ref{tab:hyperparams} summarizes the hyper-parameter setting for our experiments. In our implementation, we choose the threshold as 0.7 for percentile ranking.

\begin{table*}[t]
    \centering
    \small
    \caption{Hyper-parameter settings in our experiments.}
    \begin{tabular}{@{}cl@{}}
        \toprule
        \multicolumn{2}{l}{\textbf{Training Phase}} \\ 
        \midrule
        Offline Learning Rate & $1.414\times10^{-4}$ \\
        Online Learning Rate  & $10^{-5}$ \\
        \multirow{2}{*}{Batch Size}
            & assistant: $16$ \\
            & summary: $2$ \\
        \multirow{8}{*}{Online Generation kwargs} 
            & max\_new\_tokens (assistant): $128$ \\[0.5ex]
            & max\_new\_tokens (summary): $48$ \\[0.5ex]
            & num\_of\_samples: $1000$ \\
            & min\_length: $-1$ \\
            & top\_k: $0$ \\
            & top\_p: $0.9$ \\
            & temperature: $0.9$ \\
            & do\_sample: True \\
            & begin\_suppress\_tokens: [\texttt{tokenizer.eos\_token\_id}] \\
        \midrule
        \multicolumn{2}{l}{\textbf{Evaluation Phase}} \\ 
        \midrule
        \multirow{6}{*}{Generation kwargs} 
            & max\_new\_tokens (assistant): $128$ \\[0.5ex]
            & max\_new\_tokens (summary): $48$ \\[0.5ex]
            & min\_length: $-1$ \\
            & top\_k: $0.0$ \\
            & top\_p: $0.9$ \\
            & do\_sample: True \\
        \bottomrule
    \end{tabular}
    \label{tab:hyperparams}
\end{table*}

\section{Additional Experiment Results}
\label{app:additional_exp_results}

\subsection{Consistency in Inference Alignment}
\label{subsec:exp_4}

\begin{figure*}[t]
    \centering
    \includegraphics[width=\linewidth]{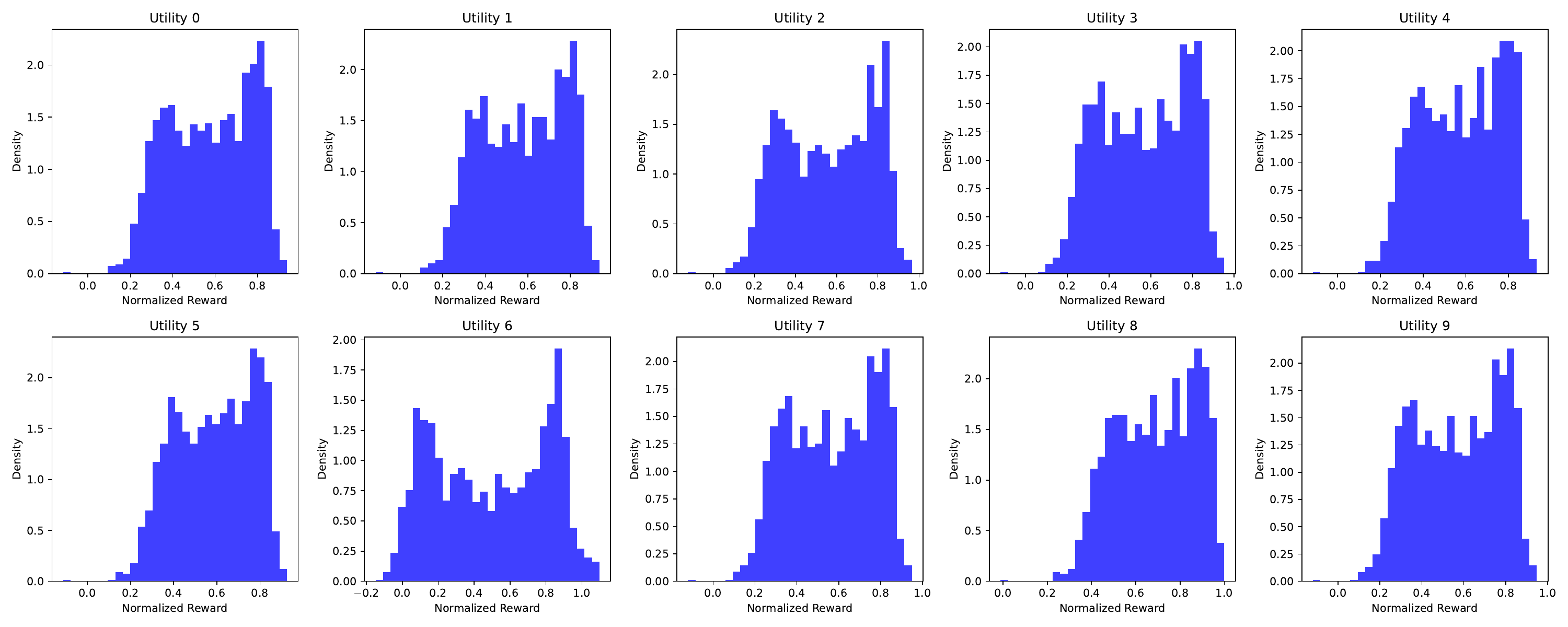}
    \caption{Utility value distribution for the ``harmless'' vs. ``humor'' task. The required utility index is 8 (i.e., the 9th utility function).}
    \label{fig:appendix_utility_idx}
\end{figure*}

We examine how closely our model’s generated responses align with a user-specified target objective, in this case, the 9th utility function (index 8). Figure~\ref{fig:appendix_utility_idx} shows the distribution of normalized utility values across all ten reward models for the ``harmless'' vs. ``humor'' task, and Table~\ref{tab:mean-norm-utility} reports the corresponding mean normalized values. Notably, the 9th utility function (index 8) attains the highest average score of 0.68, demonstrating that the LLM effectively prioritizes the user-indicated objective. In other words, when prompts include the requirement \texttt{<max\_utility\_index> i} specifying the 9th utility, the model consistently generates responses that yield higher scores along that dimension than along the others. These findings confirm that our fine-tuned model can reliably follow the user-designated target objective during inference.

\begin{table*}[t]
\centering
\small
\caption{Mean normalized value for each utility function.}
\begin{tabular}{cccccccccccc}
\toprule
\textbf{Utility index} & \textbf{0} & \textbf{1} & \textbf{2} & \textbf{3} & \textbf{4} & \textbf{5} & \textbf{6} & \textbf{7} & \textbf{8} & \textbf{9} \\
\midrule
\textbf{Mean} & 0.57 & 0.57 & 0.56 & 0.56 & 0.57 & 0.57 & 0.49 & 0.55 & \bf{0.68} & 0.58 \\
\bottomrule
\end{tabular}
\label{tab:mean-norm-utility}
\end{table*}

\subsection{Pareto Front Analysis: Four Other Tasks}
\label{subsec:exp_1}

\begin{figure*}[t]
    \centering
    \includegraphics[width=\linewidth]{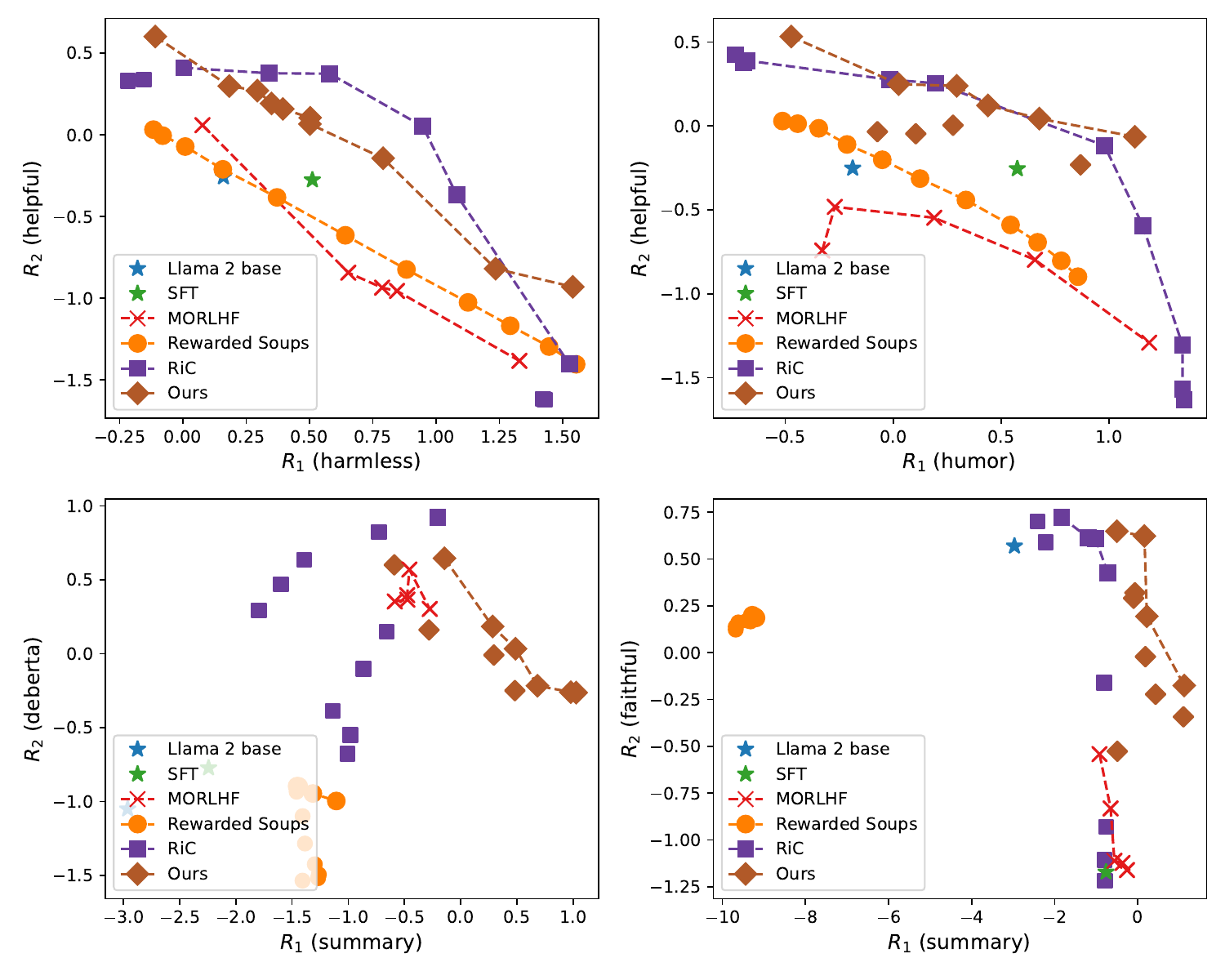}
    \caption{Pareto fronts of the remaining four tasks.}
    \label{fig:appendix_pareto_front}
\end{figure*}

In addition to the main results presented in the paper, Figure~\ref{fig:appendix_pareto_front}  shows the Pareto fronts for the remaining four tasks in our experiments. We compare five methods: Llama 2 base, SFT, Rewarded Soups, RiC, and our proposed approach (Ours). Each subplot corresponds to a different two-objective trade-off, where the $x$- and $y$-axes represent the average test rewards for two distinct objectives. A method demonstrating an \emph{outer} Pareto curve indicates a more favorable performance, as it captures a broader spectrum of trade-offs between competing objectives. Across all four subplots in Figure~\ref{fig:appendix_pareto_front}, Our method (purple line) achieves the most outward Pareto front, reflecting superior coverage of the objective space compared to the other baselines. RiC (orange line) also performs strongly but is dominated by Ours in several regions of the objective space. Meanwhile, Rewarded Soups (red line), SFT (green crosses), and Llama 2 base (blue triangles) display varying degrees of suboptimality in balancing the two objectives simultaneously. Overall, these results reinforce our main findings, showing that our method provides a robust and effective means of navigating multi-objective trade-offs across a range of different tasks.

\subsection{Ablation Study}
\label{app:ablation_study}
We conduct an ablation study to investigate how other factors (e.g., alternative design choices, and hyperparameters) influence the performance of our method.

\begin{figure*}[t]
    \centering
    \includegraphics[width=\linewidth]{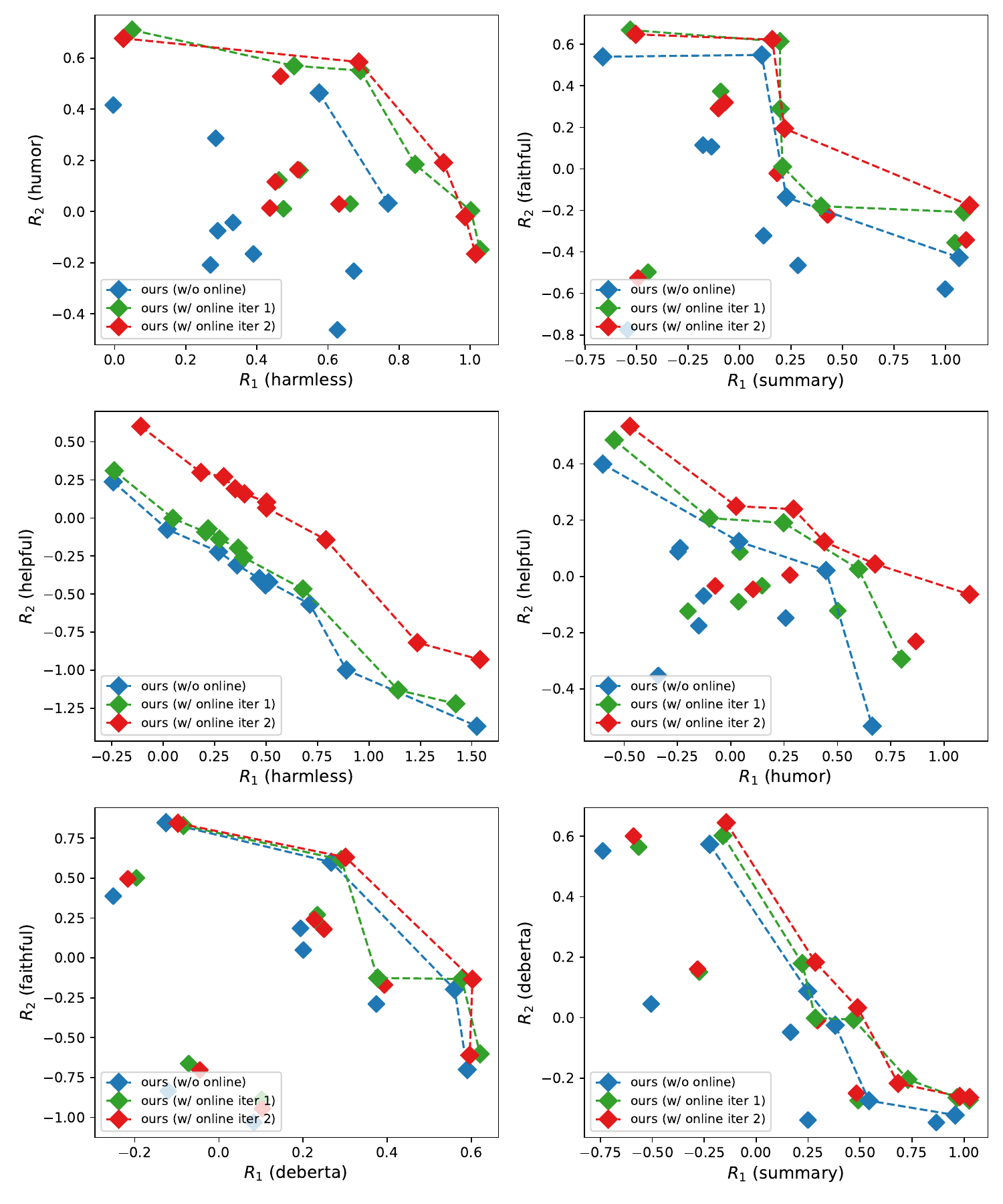}
    \caption{Ablation study of the online stage on all six tasks.}
    \label{fig:appendix_online_iter}
\end{figure*}

\noindent\textbf{Necessity of the online stage. } We perform an ablation study to compare the performance of pure offline training (``w/o online''), a single
online iteration (``w/ online iter 1''), and two online iterations (``w/ online iter 2'') across all tasks. We present the performance results for ``harmless'' vs. ``humor'' and ``summary'' vs. ``faithful'' tasks in Figure~\ref{fig:appendix_online_iter}. We observe that models incorporating online iterations consistently outperform the offline-only baseline. Notably, even a single online iteration yields significant improvements, with further performance gains observed when two iterations are applied, thereby underscoring the necessity of the online stage in our training framework. The online stage is important because it enables the model to generalize over unseen samples in the offline dataset, and ultimately enhancing its ability to generate appropriate responses corresponding to the required utility function index.

\begin{figure}[t]
    \centering
    \includegraphics[width=\linewidth]{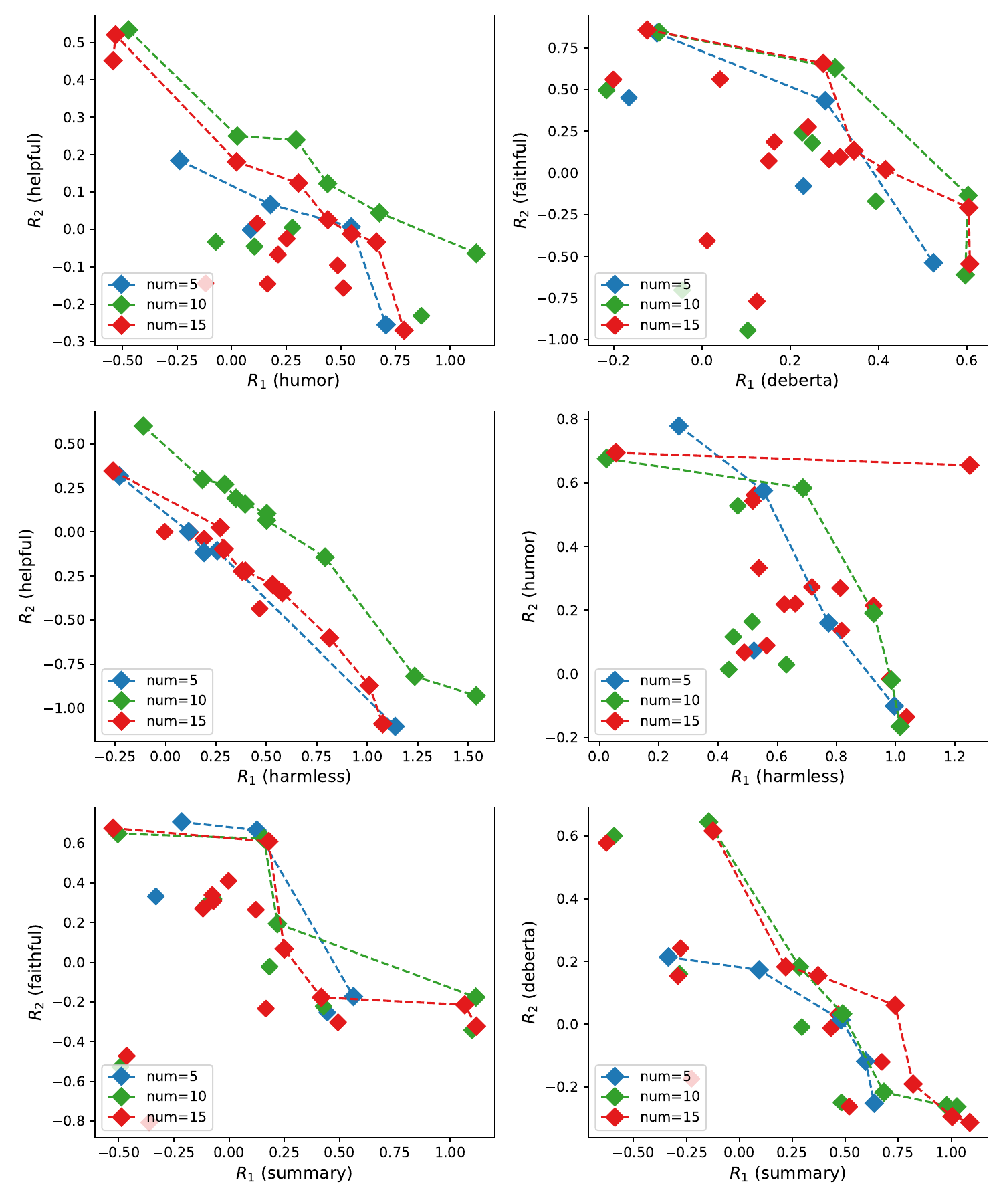}
    \caption{Ablation study of different numbers of utility functions on all six tasks.}
    \label{fig:appendix_utility_num}
\end{figure}

\noindent\textbf{Number of utility functions.}
We vary the number of utility functions from \{5,10,15\} and compare the resulting empirical Pareto fronts in Figure~\ref{fig:appendix_utility_num}. With 5 utility functions, the generated Pareto front is slightly inferior to those produced with 10 and 15 utility functions. Specifically, the front with 5 utility functions captures a narrower range of optimal trade-offs across the objectives. However, when the number of utility functions is increased to 10 or 15, the resulting Pareto fronts exhibit comparable performance. Increasing the number of utility functions beyond 10 does not yield substantial gains.
These results demonstrate that our method is insensitive to the precise number of utility functions, provided that a sufficient number (around 10) is used to effectively cover the preference space. In practical terms, employing 10 utility functions strikes a good balance between performance and computational efficiency, as further increases to 15 do not produce notable improvements.

\begin{figure*}[t]
    \centering
    \includegraphics[width=\linewidth]{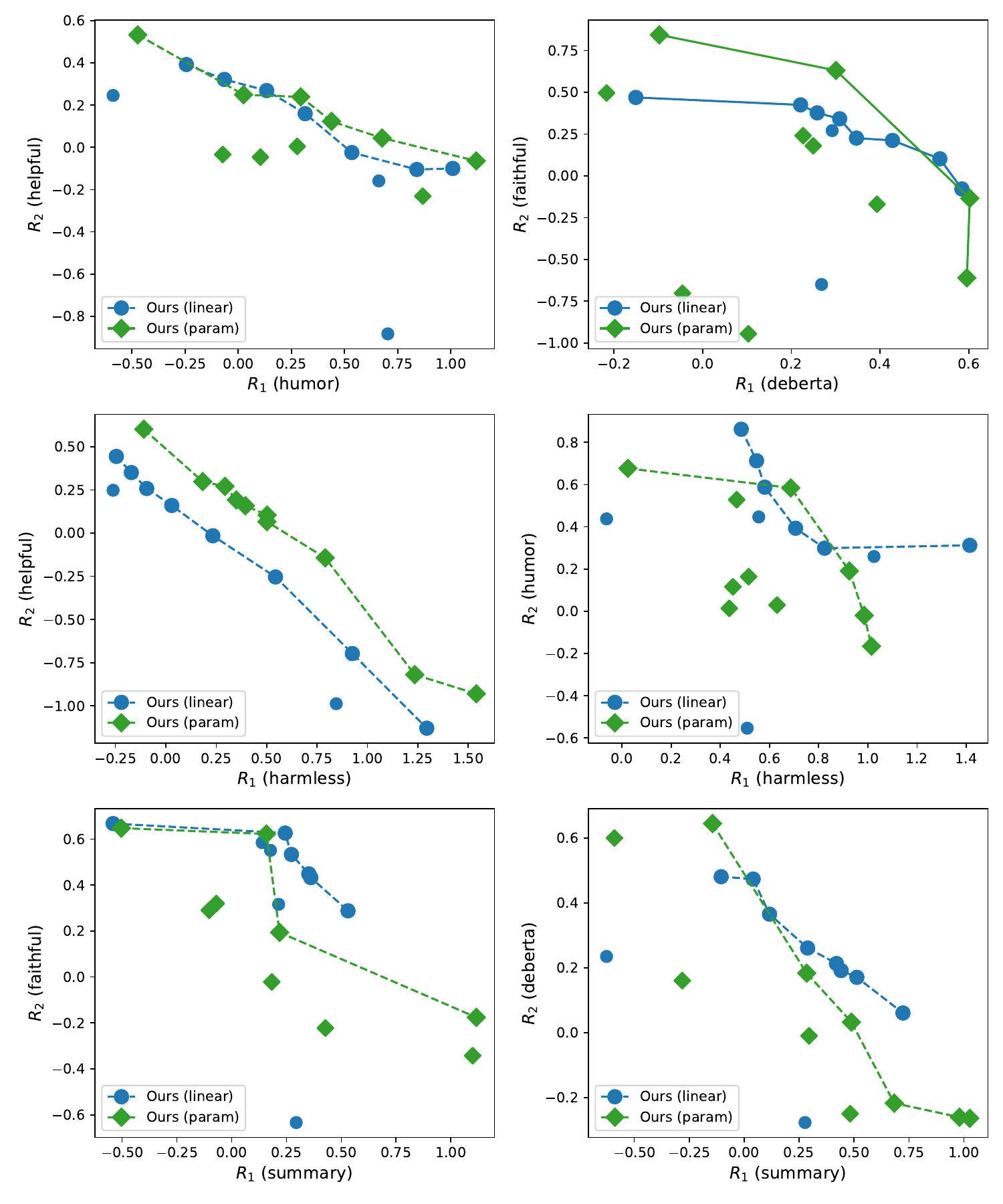}
    \caption{Abalation study of using linear utility functions.}
    \label{fig:appendix_param_linear}
\end{figure*}

\noindent\textbf{Variants of utility functions.} We investigate an alternative design choice of the utility functions, i.e., linear functions. The linear function is designed in the form of $w^T x$ where $||w||_2 = 1$. To construct $N$ linear utility functions, we uniformly sample $N$ points from the unit circle and set $w$ accordingly. We plot the Pareto fronts using linear utility functions and our method across all tasks. The generated Pareto fronts, as shown in  Figure~\ref{fig:appendix_param_linear}, illustrate that our method continues to deliver strong results even with linear utility functions. However, there is still a performance gap between the results achieved with distributional Pareto-optimal utility functions and those obtained using linear functions. While linear functions still produce reasonably good trade-offs, they fail to capture the full range of optimal solutions attainable by our method, which leverages distributional utility representations. This highlights the advantage of our approach in handling more complex utility functions, demonstrating its superior flexibility in navigating multi-objective optimization.

\section{Proof}
\label{app:proof}

The theorem in the main paper is restated here:

\setcounter{theorem}{0}
\begin{theorem}[Utility-Conditional Distributional Pareto-Optimality]
\label{thm:dp_opt_unified}
Under Assumption~\ref{ass:util_opt}, for any generated response $y \sim \pi_{\theta,i}(x)$ with reward vector $\mathbf{z}(y)$, let
\[
i^\star = \arg\max_{j \in \{1,\dots,M\}} g_{\psi_j}(\mathbf{z}(y)).
\]
Then the sub-policy $\pi_{\theta,i^\star}$ is distributionally Pareto-optimal.
\end{theorem}

We first define the First-Order Stochastic Dominance (FOSD) for distributions and policies:

\begin{definition}[FOSD for distributions]
\label{def:FOSD-distribution-unified}
Let $\boldsymbol{\mu}_1,\boldsymbol{\mu}_2$ be probability measures on $\mathcal{Z}=[0,1]^K$. We say that $\boldsymbol{\mu}_1$ \emph{first-order stochastically dominates} $\boldsymbol{\mu}_2$, written $\boldsymbol{\mu}_1 \succ_{\mathrm{SD}} \boldsymbol{\mu}_2$, iff
\[
\forall\, g_{\psi_i} \in \mathcal{U}^{\uparrow},\quad \mathbb{E}_{\mathbf{z} \sim \boldsymbol{\mu}_1}[g_{\psi_i}(\mathbf{z})] \geq \mathbb{E}_{\mathbf{z} \sim \boldsymbol{\mu}_2}[g_{\psi_i}(\mathbf{z})].
\]
\end{definition}

\begin{definition}[FOSD for policies]
\label{def:FOSD-policy-unified}
For any policy $\pi$, let $\boldsymbol{\mu}(\pi)$ denote the return distribution induced by $\pi$. Policies $\pi_1$ and $\pi_2$ are ordered by FOSD according to Definition~\ref{def:FOSD-distribution-unified}: $\pi_1 \succ_{\mathrm{SD}} \pi_2$ iff $\boldsymbol{\mu}(\pi_1) \succ_{\mathrm{SD}} \boldsymbol{\mu}(\pi_2)$.
\end{definition}

With the definition of FOSD, we have the following lemma:

\begin{lemma}[Expected-utility separation under FOSD]
\label{lem:utility-monotone-unified}
Let $u: \mathcal{Z} \to \mathbb{R}$ be \emph{strictly} increasing in each coordinate. If $\boldsymbol{\mu}_1 \succ_{\mathrm{SD}} \boldsymbol{\mu}_2$ then
\[
\mathbb{E}_{\mathbf{z} \sim \boldsymbol{\mu}_1}[u(\mathbf{z})] > \mathbb{E}_{\mathbf{z} \sim \boldsymbol{\mu}_2}[u(\mathbf{z})].
\]
\end{lemma}

\begin{proof}
By Definition~\ref{def:FOSD-policy-unified} the expected utilities are
weakly ordered for \emph{all} non‑decreasing $u$ and strictly ordered for
at least one.
Because $u$ is strictly increasing, it belongs to the latter subset,
hence the inequality is strict.
\end{proof}

Now we provide the detailed proof of Theorem~\ref{thm:dp_opt_unified}:

\begin{proof}
We aim to show that the sub-policy $\pi_{\theta,i^\star}$ is distributionally Pareto-optimal. By definition, a policy is distributionally Pareto-optimal if no other policy stochastically dominates it.

Assume, for the sake of contradiction, that there exists another policy $\pi'$ such that $\pi'$ stochastically dominates $\pi_{\theta,i^\star}$, i.e., $\pi' \succ_{\mathrm{SD}} \pi_{\theta,i^\star}$.

According to Definition~\ref{def:FOSD-policy-unified}, this implies that the return distribution of $\pi'$, denoted by $\boldsymbol{\mu}(\pi')$, first-order stochastically dominates the return distribution of $\pi_{\theta,i^\star}$, denoted by $\boldsymbol{\mu}(\pi_{\theta,i^\star})$:
\[
\boldsymbol{\mu}(\pi') \succ_{\mathrm{SD}} \boldsymbol{\mu}(\pi_{\theta,i^\star}).
\]

By Definition~\ref{def:FOSD-distribution-unified}, this means that for all non-decreasing utility functions $u \in \mathcal{U}^{\uparrow}$, we have:
\[
\mathbb{E}_{\mathbf{z} \sim \boldsymbol{\mu}(\pi')}[u(\mathbf{z})] \geq \mathbb{E}_{\mathbf{z} \sim \boldsymbol{\mu}(\pi_{\theta,i^\star})}[u(\mathbf{z})],
\]
and the inequality is strict for at least one such utility function.

Now, consider the specific utility function $g_{\psi_{i^\star}}$. From Section~\ref{subsec:utility_library} of the main paper, we know that $g_{\psi_{i^\star}}$ is strictly increasing in each coordinate (due to the added linear term with $\epsilon > 0$). Therefore, we can apply Lemma~\ref{lem:utility-monotone-unified} with $u = g_{\psi_{i^\star}}$, $\boldsymbol{\mu}_1 = \boldsymbol{\mu}(\pi')$, and $\boldsymbol{\mu}_2 = \boldsymbol{\mu}(\pi_{\theta,i^\star})$. This gives us:
\[
\mathbb{E}_{\mathbf{z} \sim \boldsymbol{\mu}(\pi')}[g_{\psi_{i^\star}}(\mathbf{z})] > \mathbb{E}_{\mathbf{z} \sim \boldsymbol{\mu}(\pi_{\theta,i^\star})}[g_{\psi_{i^\star}}(\mathbf{z})]. \quad (*)
\]

However, by Assumption~\ref{ass:util_opt}, the sub-policy $\pi_{\theta,i^\star}$ is defined as a policy that maximizes the expected value of the utility function $g_{\psi_{i^\star}}$. This means that for any other policy $\pi'$, including the one we assumed to stochastically dominate $\pi_{\theta,i^\star}$, the following must hold:
\[
\mathbb{E}_{\mathbf{z} \sim \boldsymbol{\mu}(\pi_{\theta,i^\star})}[g_{\psi_{i^\star}}(\mathbf{z})] \geq \mathbb{E}_{\mathbf{z} \sim \boldsymbol{\mu}(\pi')}[g_{\psi_{i^\star}}(\mathbf{z})].
\]

This inequality directly contradicts the strict inequality we derived from the assumption that $\pi'$ stochastically dominates $\pi_{\theta,i^\star}$. Therefore, our initial assumption that such a policy $\pi'$ exists must be false.

Since no other policy stochastically dominates $\pi_{\theta,i^\star}$, by the definition of distributional Pareto-optimality, the sub-policy $\pi_{\theta,i^\star}$ is indeed distributionally Pareto-optimal.
\end{proof}

\section{User Study}
\label{app:user_study}

To evaluate the qualitative performance of UC-MOA compared to the state-of-the-art method RiC, we conducted a user study where participants rated anonymized responses from both systems to a series of diverse prompts. This section details the design of the study, the presentation of prompts and system responses, and an analysis of the collected user preferences and qualitative feedback.

\subsection{User Study Design}
\label{app:user_study_design}

\textbf{Objective:} The primary goal of this study was to compare user preferences for responses generated by UC-MOA versus RiC across a range of tasks requiring a blend of humor, information, and practicality.

\textbf{Participants:} We utilized the criterion sampling method~\cite{palinkas2015purposeful} to recruit participants. The participants were expected to have English reading skills and hold some basic knowledge of large language models.

\textbf{Task:} For each of the five prompts, participants were presented with two anonymized system responses. The assignment of responses to "System A" and "System B" was randomly permuted for each prompt and each participant to prevent bias. Participants were not informed which system corresponded to UC-MOA or RiC. Participants were asked to choose which of the two responses they preferred and were given the option to provide a brief written justification for their choice.

\subsubsection{Participant Instructions}
\label{app:participant_instructions}

We first present the following general instructions:

Welcome to our study on evaluating AI-generated text. Your task is to help us understand which responses people find more humor. We will present 5 questions below. For each question, you will see a prompt that was given to two different AI systems. Below the prompt, you will see two responses, labeled ``System A'' and ``System B''. These labels are randomized. Please read both responses carefully and select the response (System A or System B) that you prefer. Optionally, you can provide a brief explanation for your choice in the text box provided. Your feedback is valuable in helping us understand your preferences. There are no right or wrong answers; we are interested in your genuine preference. Thank you for your participation!

Here are five prompts and the corresponding responses:
\begin{enumerate}
\item \emph{Explain quantum computing to a 12--year--old in a funny way.}
\begin{itemize}
\item \textbf{System~A}: Quantum computing is like giving your homework to a coin that can be both heads and tails at once---and the coin roasts you about Schr\"{o}dinger's cat while it solves algebra!
    \item \textbf{System~B}: Imagine your game console pressing every button at the same time to find the winning combo, all while cracking puns about super positions.
\end{itemize}

\item \emph{I'm anxious about my job interview tomorrow. Give me three concrete tips that also lighten the mood.}
\begin{itemize}
\item \textbf{System~A}: 1.~Breathe like Darth~Vader on a beach holiday---slooow in, slooow out. 2.~Write your top strengths on a sticky note and fist\-bump it like a tiny hype\-buddy.
    \item \textbf{System~B}: Picture your nerves as a mischievous house-cat: acknowledge it, pet it once, then move it off your keyboard so you can prep.
\end{itemize}

\item \emph{Write a short, humorous yet informative comparison between cats and dogs as pets.}
\begin{itemize}
\item \textbf{System~A}: Cats are the introverted artists of the animal kingdom: low\-maintenance, occasionally aloof, and prone to 3~a.m.~parkour. Dogs are the extroverted linebackers: loyalty dialed up to~11, endless fetch, and zero respect for personal space.
    \item \textbf{System~B}: Dog owners get free cardio sessions (walkies!), while cat owners get free zen lessons (meditative ignoring). Both come with bonus fur-glitter for your clothes.
\end{itemize}

\item \emph{Help me troubleshoot why my sourdough starter isn't rising, with a dash of comedy.}
\begin{itemize}
\item \textbf{System~A}: If your starter is lazier than a teenager on a Sunday, first make sure it's actually getting food: equal\-weight flour and water, twice a day. Think of feeding it as paying rent to billions of yeasty tenants.
    \item \textbf{System~B}: Pretend your starter is a micro-pet dragon: it needs warmth (25	extdegree{}C), snacks (carbs), and breathing room. Starve it or freeze it and you'll have a very sleepy dragon.
\end{itemize}

\item \emph{Draft a witty but practical email reminding my roommate to pay their share of the rent.}
\begin{itemize}
\item \textbf{System~A}: \texttt{Subject: Friendly Rent Reminder}\newline Hey~[Roomie], just popping into your inbox like the first\-of\-the\-month fairy. The rent goblin is hungry for your share (\$650) by Friday.
    \item \textbf{System~B}: \texttt{Subject: The First of the Month Strikes Again}\newline Dearest co-dweller, our beloved landlord prefers money to memes, and the ritual tribute (\$650) falls due this week.
\end{itemize}
\end{enumerate}

All system outputs are truncated to the first two sentences for brevity.

\subsection{Representative Participant Comments}
\label{app:prompts_responses_feedback}
Participants were invited (optionally) to write a brief justification after each vote. Below are representative, lightly edited remarks that capture common themes.

\begin{itemize}
\item \textbf{P1}: \emph{System~A explained the idea more clearly while still making me laugh---System~B's joke landed but felt vague.}
  \item \textbf{P2}: \emph{A's breathing tip and sticky-note trick were immediately actionable; B's cat metaphor was cute but less useful.}
\item \textbf{P3}: \emph{I liked A's vivid imagery of 3~a.m.~parkour; B was funny too but didn't contrast the pets as sharply.}
  \item \textbf{P4}: \emph{System~B's dragon analogy was hilarious, but A actually told me what temperature to use---so I picked A.}
\item \textbf{P5}: \emph{Both emails were witty, but A gave a concrete amount and deadline, which felt more practical.}
\end{itemize}

These qualitative comments mirror the quantitative outcome: participants consistently favored answers that blended humor with concrete, actionable advice---qualities more frequently exhibited by UC-MOA outputs.

\subsection{Overall Findings and Discussion}
\label{app:overall_findings}

The user study results, encompassing both the quantitative preferences (46 out of 60 favor our method while only 14 out of 60 favor RiC) and the qualitative participants' comments, reveal a consistent pattern. Participants generally favored responses that effectively balanced humor with clear, concrete, and actionable information. The qualitative feedback, as exemplified by the comments in Section~\ref{app:prompts_responses_feedback}, indicates that while both systems were capable of generating humorous or engaging text, the perceived utility and directness of the advice often served as key differentiators in user preference. These attributes were more consistently observed in the outputs generated by UC-MOA, contributing to its higher overall preference rate in this study. This suggests that UC-MOA demonstrates a stronger ability to integrate an engaging style with practical substance, thereby aligning more closely with user expectations for the types of prompts investigated.

\end{document}